\documentclass{article}
\usepackage{proceedings}
\usepackage[latin1]{inputenc}
\usepackage{graphicx}
\usepackage{amssymb}

\newtheorem{Def}{Definition}
\newtheorem{Ex}{Example}

\newtheorem{Pro}{Property}

\newcommand{\abo}{{\mathcal{A}}}
\newcommand{\fbo}{{\mathcal{F}}}
\newcommand{\doadmissible}{restrictedly admissible}
\newcommand{\doadmissibility}{restricted admissibility}

\newtheorem{Demonstration}{Proof}

\newenvironment{Demo}{\begin{Demonstration}}{$\diamondsuit$\end{Demonstration}}

\begin{document}

\title{``Minimal defence'': a refinement of the preferred semantics for
   argumentation frameworks}

\author{ {\bf C. Cayrol} \\
IRIT - UPS \\
118, rte de Narbonne\\
31062 Toulouse, France \\
\{{\tt ccayrol@irit.fr}\} \\
\And
{\bf S. Doutre} \\
IRIT - UPS \\
118, rte de Narbonne\\
31062 Toulouse, France \\
\{{\tt doutre@irit.fr}\}\\
\And
{\bf M.-C. Lagasquie-Schiex} \\
IRIT - UPS \\
118, rte de Narbonne\\
31062 Toulouse, France \\
\{{\tt lagasq@irit.fr}\}\\
\And
{\bf J. Mengin} \\
IRIT - UPS \\
118, rte de Narbonne\\
31062 Toulouse, France \\
\{{\tt mengin@irit.fr}\} \\
}

\maketitle

\begin{abstract}
Dung's abstract framework for argumentation enables a study of the
interactions between arguments based solely on an ``attack'' binary relation
on the set of arguments. Various ways to solve conflicts between
contradictory pieces of information have been proposed in the
context of argumentation, nonmonotonic reasoning or logic programming,
and can be captured by appropriate semantics within Dung's framework.
A common feature of these semantics is that one can always maximize in
some sense the set of acceptable arguments. We propose in this paper to
extend Dung's framework in order to allow for the representation of what
we call ``restricted'' arguments: these arguments
should only be used if absolutely necessary, that is,
in order to support other arguments that would
otherwise be defeated. We modify Dung's preferred
semantics accordingly: a set of arguments becomes
acceptable only if it contains a minimum of restricted
arguments, for a maximum of unrestricted arguments.
\end{abstract}

\section{INTRODUCTION}\label{intro}

Argumentation is a form of reasoning that has recently captured the
interest of many researchers in the Artificial Intelligence community,
among others. An abstract framework for studying it has been proposed by
Dung \cite{Dun95}. It is based on the assumption that the process of
argumentation is built from two types of jobs: the formation of
arguments, associated with the interactions that exist between them; and
the evaluation of the validity of the arguments. A form of interaction
between arguments that is often difficult to ignore is contradiction,
since a usual basic measure of the validity of a piece of reasoning is
the level of self-contradiction it contains. In Dung's approach, an
argumentation framework consists of two elements: a set of arguments, and
a binary relation between arguments, so that one can represent an
``attack'', or contradiction, relation between arguments. Thus, the set
of arguments and the attack relation form a graph.

\medskip

If one accepts that the graph contains all needed information, it is
possible to define which arguments are valid, or acceptable, by
considering solely this graph. Our intuition tells us that an argument
which is not attacked at all should be acceptable. More generally, one can
define that an acceptable argument is one which is defended, that is,
whose attackers are all attacked by arguments which are themselves
acceptable.

When the graph contains cycles, this definition is circular, and one
needs a more global approach. A solution is to define acceptable sets of
arguments. Arguments in such a set do usually not attack each other, and
defend each other: every attacker of an argument in the set is itself
attacked by an argument in the set. This definition leads to Dung's
``admissible'' sets of arguments. Maximal sets of arguments that are
acceptable are then called ``preferred extensions'' of the framework.

\medskip

These definitions of acceptability make no distinction between
arguments, in the sense that defenders are all equally desirable: if an
argument can be independently defended by two different arguments, then
one may choose any of the two, or even the two arguments together if
they do not attack each other, to defend the first one. However, there
are problems where a distinction between several levels of desirability
of defenders is needed. Suppose for example that our arguments
``belong'' to two agents: one, the proponent, is trying to make a point,
while the other one, the opponent, is trying to refute it. The proponent
may have arguments that should only be disclosed if absolutely necessary
(like very personal information).

We propose below to refine Dung's argumentation framework with a
distinction between two types of arguments: we consider \emph{unrestricted}
arguments, that one can maximize, and \emph{restricted} arguments, the use of
which will be kept to a minimum. Restricted arguments should only be used
if absolutely necessary, that is, in order to support other arguments
that would otherwise be defeated.

Notice that the distinction between restricted and unrestricted arguments
is at a defensive level. Other frameworks have been designed in order
to authorize varying strengths for the arguments. In the frameworks
described in \cite{AmgoudCayrol:98} and \cite{PrakkenSator:97}, a preference
relation among arguments is taken into account when the interactions
between arguments are studied: the idea is that in order to be a
serious menace, an argument must be preferred to the argument it
attacks. In this case, there is no need to minimize the number of
arguments of a given strength, except maybe for a purpose of
conciseness or of efficiency.

\medskip

We also propose to allow for a distinction between the arguments of the
proponent and those of the opponent: this enables us
to ensure, when this is needed, that the arguments of the opponent are
never used to defend an argument of the proponent. Note that this
distinction is independent from the previous one, between restricted
and unrestricted arguments.

\medskip

The paper is built as follows: our refined argumentation framework is
presented in the next section. In particular, we detail the partition
of the set of arguments, and provide a new definition for
admissibility; we then define a new type of extensions for our refined
argumentation framework. We illustrate our definitions with three
particular argumentation frameworks in section~\ref{examples}. 
Finally, we conclude
and sketch future works in section~\ref{algorithms}.

\section{FORMALIZATION} \label{formalization}

We recall the definition of \cite{Dun95}'s argumentation framework.

\begin{Def}
\cite{Dun95}
An {\em argumentation framework} is a pair $\abo\fbo= (A, R)$ where 
$A$ is a set of arguments
and $R$ is an attack relation between arguments ($R \subseteq A \times
A$).

An argumentation framework is {\em well-founded} if and only if there is
no infinite sequence

\centerline{$a_0$,  $a_1$,  \ldots,  $a_n$,  \ldots}

such that $\forall i,  a_i \in A$ and $a_{i+1} R a_i$\footnote{If $R$ 
is finite, an
argumentation framework  is well-founded if and only if it does not 
contain any cycle. }.
\end{Def}

An argumentation framework can  easily be represented as a directed
graph, where vertices are the arguments and edges correspond to the
elements of the relation $R$.

\begin{Ex}
Let $\abo\fbo_1 = (A, R)$ be an argumentation framework such that:

$A = \{ u_1, u_2, u_3, u_4, u_5, r_1, r_2, r_3, r_4, o_1, o_2, o_3,
o_4, o_5 \}$ and

$R = \{ (o_1, u_1), (o_2, u_2), (u_3, o_2), (o_3, u_4), (u_5, o_4), 
(r_1, o_2),$ \\ $(r_2, o_3), (o_4, r_3), (o_5, r_4) \}$.

The graph representation of $\abo\fbo_1$ is depicted on 
figure~\ref{fig_example1}.

\begin{figure}[h]
   \begin{center}
     \includegraphics{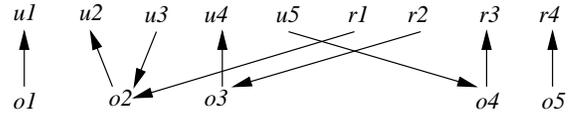}
   \end{center}
   \caption{Graph representation of the argumentation framework 
$\abo\fbo_1$} \label{fig_example1}
\end{figure}

\end{Ex}

\paragraph{Partitions of the set of arguments of an argumentation framework}
Let $\abo\fbo= (A, R)$ be an argumentation framework.
In order to refine some existing semantics for argumentation
and considering that $A$ represents the arguments of
different agents, the meaning of a partition of $A$ may be the following:

\begin{itemize}
\item A simple case is to take into account one agent against the
   others. So we have a simple partition of $A$, called
   \texttt{type-1-partition}: $A = F \cup (A \setminus F)$ for a given set
   of  arguments $F$ (corresponding to the arguments of the selected agent);

   the argumentation framework $\abo\fbo$
   {\tt type-1-} partitioned, given a set
   of arguments $F$, is denoted by $\abo\fbo_F$;

\item the other type of a partition corresponds to the distinction
   between restricted arguments (which should not be used
   if possible) and unrestricted arguments. So, we have a special 
partition of $A$,
   called \texttt{type-2-partition}, which is a refinement of a
   \texttt{type-1-partition}: given a set $F$ such that $F = F_u \cup
   F_r$ with $F_u \cap F_r = \varnothing$, $A = F_u \cup F_r \cup (A
   \setminus (F_u \cup F_r))$ ($F_u$,  $F_r$ are respectively the set 
of the unrestricted
   arguments and the set of the restricted
   arguments of the selected agent);

   the argumentation framework $\abo\fbo$ {\tt type-2-} partitioned, is
   denoted by $\abo\fbo_{F_{ru}}$.
\end{itemize}

Note that, although every \texttt{type-2-partition} has an underlying
\texttt{type-1-partition}, the latter can be a vacuous one: it is possible
to set $ F $ to be the entire set $ A $.


\subsection{ADMISSIBILITY} \label{admissibility}

Let $\abo\fbo = (A, R)$ be an argumentation framework,  we have the 
following definitions.

\begin{Def}
\cite{Dun95}
A set $S\subseteq A$ is \emph{conflict-free} if and only if $\nexists
a, b \in S$ such that $a R b$.
\end{Def}

\cite{Dun95} defines the notion of defence. Defence can be of two
kinds: collective or individual. Collective defence is achieved by a
set of arguments, individual defence by one argument.

\begin{Def}
\cite{Dun95}
Let $S \subseteq A$,  $a \in A$.  \emph{$S$ defends (collectively) $a$} if
and only if $\forall b \in A$,  if $b R a,  \exists c \in S$ such that
$c R b$.  \emph{$S$ defends all its elements} if
and only if $\forall a \in S$,  if $\exists b \in A$ such that $b R a$
then $\exists c \in S$ such that $c R b$.
\end{Def}

The following definition is inspired by~\cite{Dun95}:

\begin{Def}
Let $a,  c \in A$.  \emph{$c$ is an individual defender of $a$} if and
only if there is a finite sequence $x_0, \ldots, x_{2n}$ such that:
\begin{enumerate}
\item $a=x_0$ and $c=x_{2n}$,  and
\item $\forall i,  0 \leq i \leq (2n - 1),  x_{i+1} R x_i$.
\end{enumerate}
If $n = 1$,  $c$ is a \emph{direct individual defender} of $a$.
\end{Def}

Then,  \cite{Dun95} proposes the admissible semantics:

\begin{Def}\label{def_adm}
\cite{Dun95} A set $S \subseteq A$ is \emph{admissible} if and only
if $S$ is conflict-free and  $S$ defends all its elements.
\end{Def}

Every argumentation framework has at least one admissible set because 
the empty set is admissible.

\begin{Ex}
The sets $\emptyset$, $\{ o_1 \}$, $\{o_1, u_2, u_3 \}$, $\{ o_1, u_2,
u_3, u_4, u_5, r_1, r_2, r_3, o_5 \}$ are admissible sets of the
argumentation framework $\abo\fbo_1$ represented on
figure~\ref{fig_example1}.
\end{Ex}

We refine this admissible semantics in the context of
a \texttt{type-2-partition} of $A$ in order to accept restricted
arguments only if their presence is justified, that is when they
defend at least one unrestricted argument. This refinement is called the
{\em  \doadmissibility{}}.

Let $\abo\fbo_{F_{ru}}$ be a {\tt type-2-}partitioned argumentation
framework, we have the following notations and definition:

\paragraph{Notations}
Given $F \subseteq A$ such that $F = F_u \cup F_r$ and $F_u \cap F_r =
\varnothing$ and a set $S \subseteq A$,   $S_u$ denotes $S \cap F_u$
and $S_r$ denotes
$S\cap F_r$.  We say that $S_u$ is the  \emph{unrestricted} part of
$S$,  and $S_r$ is the \emph{restricted} part of $S$.

\begin{Def}\label{def_do_adm}
Let $S \subseteq F$.  $S$ is  \emph{\doadmissible{}} if and only if:
   \begin{enumerate}
   \item $S$ is admissible and
   \item $\forall x \in S_r$,  $\exists y \in S_u$ such that $x$
     is an individual defender of $y$.
   \end{enumerate}
\end{Def}

\begin{Ex}
The set of arguments of the argumentation framework $\abo\fbo_1$ depicted on
figure~\ref{fig_example1}, can be partitioned as follows: let $F = F_u
\cup F_r$ with $F_u = \{ u_1, u_2, u_3, u_4, u_5 \}$, $F_r = \{ r_1,
r_2, r_3, r_4 \}$ and $A \setminus F = \{ o_1, o_2, o_3, o_4, o_5
\}$. The resulting {\tt type-1-} (resp. {\tt type-2-}) partitioned 
argumentation
framework is called $\abo\fbo_2$ (resp. $\abo\fbo_3$). $\abo\fbo_2$
and $\abo\fbo_3$ are depicted on figure~\ref{fig_example2}.

\begin{figure}[h]
   \begin{center}
     \includegraphics{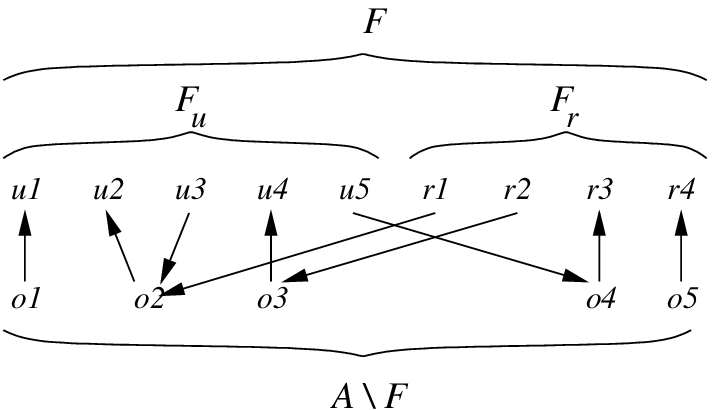}
   \end{center}
   \caption{Argumentation frameworks $\abo\fbo_2$ and  $\abo\fbo_3$} 
\label{fig_example2}
\end{figure}

$\abo\fbo_3$ has several \doadmissible{} sets. For example:

\centerline{$\emptyset$, $\{ u_2, u_3\}$,
$\{ u_2, u_3, u_4, u_5, r_2 \}$, $\{ u_2, u_3, u_4, u_5, r_1, r_2\}$.}

The set $S = \{ u_5, r_3 \}$ is not \doadmissible{}
since $r_3$ does not defend any argument of the unrestricted part of $S$
(that is $\{ u_5 \}$).
\end{Ex}


\subsection{EXTENSIONS} \label{extension}

Let $\abo\fbo = (A, R)$ be an argumentation framework. \cite{Dun95}
has also defined the preferred semantics in which an acceptable set of
arguments is called a {\em preferred extension}.

\begin{Def}\label{def_ext_pref}
\cite{Dun95} A set $S\subseteq A$ is a \emph{preferred extension} of
$\abo\fbo$ if and
only if $S$ is maximal for set-inclusion among the admissible sets of 
$\abo\fbo$.
\end{Def}

We recall:

\begin{Pro} \label{pro_adm_pref}
\cite{Dun95}
\begin{enumerate}
\item \label{pro_adm_pref1} Every admissible set of $\abo\fbo$ is
   included in a preferred extension of $\abo\fbo$.
\item Every argumentation framework has at least one preferred extension.
\item Every well-founded argumentation framework has one and only one 
preferred extension.
\end{enumerate}
\end{Pro}

\begin{Ex}
$\abo\fbo_1$, $\abo\fbo_2$ and $\abo\fbo_3$ have only one preferred
extension:

\centerline{$\{ o_1, u_2, u_3, u_4, u_5, r_1, r_2, r_3, o_5 \}$.}
\end{Ex}

In order to privilege some arguments of $A$, and to facilitate the computation
of the extensions under the new semantics, we define a {\em preferred
extension on} a given set $X$. Note that it implies the use of a {\tt
type-1-partition}  of $A$: $A = X \cup (A \setminus X)$.

\begin{Def}\label{def_ext_pref_on_a_set}
Let $\abo\fbo = (A, R)$ be an argumentation framework and let $X 
\subseteq A$.  A
\emph{preferred extension on $X$} of $\abo\fbo$ is an admissible set $S
\subseteq X$ such that $S$ is $\subseteq$-maximal on $X$,
{\it i.e.}  $\forall Y \subseteq (X \setminus S)$,  $Y \cup S$ is not 
admissible.
\end{Def}

Note that~:
\begin{itemize}
\item The preferred extensions on $A$ in the sense of
   definition~\ref{def_ext_pref_on_a_set} are the preferred extensions
   in the sense of definition~\ref{def_ext_pref}.
\item A preferred extension on $X$ is not simply the intersection of
   $X$ and of a
   preferred extension of $\abo\fbo$. For example, let $\abo\fbo = (\{a, b, c\},
   \{(c,b),(b,a)\})$ be an argumentation framework and $X = \{a\}$, the
   preferred extension of $\abo\fbo$ is $\{c, a\}$, its intersection
   with $X$ is $\{a\}$ and the only preferred extension on $X$ is 
$\varnothing$~!
\end{itemize}

The following results hold:

\begin{Pro} \label{pro_ext_def} \label{pro_pref_2}
Let $\abo\fbo = (A, R)$ be an argumentation framework and let $X\subseteq A$.
\begin{enumerate}
\item \label{pro_adm_pref_2} Every admissible set included in $X$ is
   included in a\ preferred extension on  $X$.
\item \label{pro_pref} Every preferred extension on $X$ is included in a
   preferred extension of $\abo\fbo$.
\item Let $(X, R')$ be the restriction of  $\abo\fbo$ to
   $X$\footnote{$R' = \{(a, b) | a R b \mbox{ and } a \in X,  b \in X\}$.}.
   If $(X, R')$ is a well-founded argumentation framework
   then $\abo\fbo$ has one and only one preferred extension on $X$.
\end{enumerate}
\end{Pro}

The proof of this property (and the detailed proofs of all the 
properties of this paper,
except the property~\ref{pro_adm_pref} which is given in~\cite{Dun95}) is
given in~\cite{CDLSM02}.

\begin{Ex} \label{ex_5}
$\abo\fbo_2$ has one preferred extension on $F$: $\{ u_2, u_3, u_4, 
u_5, r_1, r_2, r_3 \}$.
\end{Ex}

Preferred extensions are interesting in several respects:
\begin{itemize}
\item they provide a good ``summary'' of the admissible sets,
   since every subset of a preferred extension can be completed
   to an admissible set, and every admissible set is subset of
   at least one preferred extension;
\item the intersection of the preferred extensions can be
   interpreted as being the set of arguments that cannot be defeated.
\end{itemize}
Similar remarks can be made for other types of extensions of argumentation
frameworks, like stable extensions for example. Similarly,
we want to define a family of ``best'' \doadmissible{}
sets. In the context of a {\tt type-2-}partitioned
argumentation framework, the idea is to select the \doadmissible{} sets
which contain a minimum of restricted arguments for a maximum of 
unrestricted arguments.
In order to do so, we define the following relation:

\begin{Def}
Let $\abo\fbo_{F_{do}}$ be a
{\tt type-2-}partitioned argumentation framework. Let $S_1$ and $S_2$
be two subsets of $F$.  $S_2$ is \emph{$\prec$-better than}
$S_1$  (denoted by $S_1 \prec S_2$) if and only if $S_{1u} \subset
S_{2u}$,  or $S_{1u} = S_{2u}$ and $S_{2r} \subseteq S_{1r}$.
\end{Def}

The relation $\prec$ is a partial order on the set of the subsets of
$F$, which is clearly different from set-inclusion. The meaning of
this relation is the following: $S_1 \prec S_2$ if
and only if $S_2$ contains more unrestricted arguments than  $S_1$, or
$S_1$ and $S_2$ have the same unrestricted arguments but  $S_2$
contains less restricted arguments than $S_1$.

The idea of the new semantics is to have as few restricted
arguments  as possible in a \doadmissible{} set. In other words, the set of
restricted defenders must be minimal. That is why we call the new semantics the
\emph{minimal defence semantics}, or, for short, the \emph{min-def}
semantics. The acceptable sets under this semantics are called
\emph{min-def extensions}.

\begin{Def}\label{def_do_pref}
$S$ is a \emph{min-def extension} of $\abo\fbo_{F_{ru}}$ if and only
if $S$ is $\prec$-maximal among the \doadmissible{} sets of 
$\abo\fbo_{F_{ru}}$.
\end{Def}

Note that the restricted arguments  in an acceptable set under
the new semantics are such that, not only they defend an unrestricted
argument $x$, but they are essential defenders since there is no unrestricted
argument which defends $x$. Remark also that not every preferred extension is a
min-def extension.

\begin{Ex}
$\abo\fbo_3$ has only one min-def extension: $S = \{ u_2, u_3, u_4,
u_5, r_2 \}$ (which is different of its preferred extension). $S$ 
does not contain the
restricted argument $r_1$ whereas
$r_1$ defends $u_2$ against $o_2$. The reason is that the unrestricted
argument $u_3$ defends $u_2$, therefore $r_1$ is not useful.
\end{Ex}



We have proved a number of properties. The first one establishes the
link between the admissibility and the min-def
semantics (note that the admissible sets defined in $\abo\fbo$ are
the sames in $\abo\fbo_{F_{ru}}$):

\begin{Pro}\label{pro_resultat1}
Let $S \subseteq F$.  $S$ is a min-def extension of $\abo\fbo_{F_{ru}}$ if
and only if $S$ is  $\prec$-maximal among the admissible sets of $\abo\fbo$.
\end{Pro}
\begin{Demo} [Sketch]
It is a consequence of the fact that every admissible set which is
$\prec$-maximal is \doadmissible{}.
\end{Demo}

The following property shows that every finite
argumentation framework has at least one min-def extension:

\begin{Pro} \label{pro_existence}
Suppose that $ A $ is finite. Then for every admissible set $G$ 
included in $F$,
there is a min-def
extension $S$ of $\abo\fbo_{F_{ru}}$ such that $G \prec S$.
\end{Pro}
\begin{Demo} [Sketch]
Given $Adm_F$ the set of the admissible sets of $\abo\fbo$ included in
$F$ and the relation $\prec$, we prove that every chain in $Adm_F$ 
has an upper bound in
$Adm_F$.
\end{Demo}

The next two properties explain the link between preferred extensions
on a given set and min-def extensions. This link is very interesting
in a computational perspective (see section~\ref{algorithms}).

\begin{Pro}\label{pro_consequence1}
If $S$ is a min-def extension of $\abo\fbo_{F_{ru}}$,  then there is a
preferred extension  $E$ on $F$ such that $E_u = S_u$ and $E_r \supseteq
S_r$,  so that $E \prec S$.
\end{Pro}
\begin{Demo} [Sketch]
The proof relies upon property~\ref{pro_resultat1} and the
following result:

If $G$ is a  $\prec$-maximal admissible set included
in $F$ and if $H$ is an admissible set which contains $G$, then $H_u =
G_u$.

Since every admissible set included in $F$ is included in a preferred
extension on $F$ (see
property~\ref{pro_pref_2}), there exists a
preferred extension $E$ on $F$ such that $G \subseteq E$ and then $G_u = E_u$.
\end{Demo}

\begin{Pro}\label{pro_consequence2}
Let $E$ be a preferred extension on $F$.  There
is a min-def extension $S$ of $\abo\fbo_{F_{ru}}$ such that $E_u
\subseteq S_u$.
\end{Pro}
\begin{Demo} [Sketch]
We show that if $G$ is an admissible set included in $F$, then there
exists an admissible set $H$ included
in $F$ maximal for the relation $\prec$ such that $G_u \subseteq
H_u$. The property is a consequence of this result and of
property~\ref{pro_resultat1}.
\end{Demo}

It can be noticed that the preferred extension on $F$ of
example~\ref{ex_5} is such that its unrestricted part $\{ u_2, u_3, u_4,
u_5 \}$ is equal to the unrestricted part of the min-def extension of
$\abo\fbo_3$, and its restricted part $\{ r_1, r_2, r_3 \}$ contains the
restricted part $\{ r_2 \}$ of the min-def extension of $\abo\fbo_3$.

\section{ILLUSTRATIONS} \label{examples}

Let us give three illustrations of the formal framework presented in 
section~\ref{formalization}.
The first two are examples in a dialogical context, and the last one 
shows that the framework
can be applied to other contexts.

\subsection{A DISCUSSION BETWEEN FRIENDS}

Denis and Theo have a discussion. Denis has a point of view and gives
arguments to support it. Denis has a friend, Olivia, who agrees with
him. Theo does not agree with Denis and Olivia's point of view and
gives arguments attacking their arguments. In this discussion, Denis
can be viewed as the proponent and Theo as the opponent. Denis has
several solutions to defend his arguments against Theo's attacks:
\begin{enumerate}
\item \label{cas1}either Denis uses his own arguments only;
\item \label{cas2}or Denis also accepts Olivia's arguments and
   considers that Olivia's arguments have the same importance as his:
   he uses indiscriminately his arguments or hers to defend his
   arguments;
\item \label{cas3}or Denis accepts Olivia's arguments but he wants to
   privilege his own arguments: he takes into account Olivia's
   arguments only when they defend some of his arguments which he
   cannot defend with his own arguments.
\end{enumerate}

The different ways Denis can accept Olivia's arguments are justified by
the fact that Olivia's argument could contain, for example, very
personal information on Denis. Therefore, one can imagine that if
Denis does not really want these informations to be revealed, he
accepts Olivia's arguments only if absolutely necessary
(case~\ref{cas3}) or he does not accept them at all
(case~\ref{cas1}). On the contrary, if Denis agrees to reveal these
informations, he makes no distinction between Olivia's arguments and
his (case~\ref{cas2}).

\medskip

The arguments and the attacks of the
discussion constitute an argumentation framework where Denis'
arguments are the unrestricted arguments, Olivia's arguments are the
restricted arguments and Theo's arguments are the other arguments.

We assume that the acceptable sets of arguments supporting Denis'
point of view are conflict-free sets which defend all their arguments
against Theo's attacks and contain a maximum of Denis'
arguments. Thus, in case~\ref{cas1}, these sets are preferred
extensions on Denis' arguments; in case~\ref{cas2}, they are
\doadmissible{} sets; in case~\ref{cas3}, they are min-def extensions.

\medskip

In a multi-agent framework, it is usually the case that the agents
need to communicate to fulfil tasks or to share
resources, so that they have to engage in dialogues. The
works of \cite{PSJ98}, \cite{Ree98} and \cite{AMP00} outline that argumentation
can be used as a basis for dialogues. \cite{AMP00} studies different
types of dialogue in argumentation theory, and mentions the strategies
that an agent uses to choose an argument during a dialogue. This 
point is more precisely
investigated in \cite{AM02}. The example above suggests that our work 
could be used as a
support to implement the strategies. The unrestricted arguments of an 
agent would be the ones
which satisfy the strategy, the restricted arguments would be the 
other arguments of the agent.


\subsection{MAFIA ON TRIAL}

Consider a criminal court case with the mafia on trial, and
assume that some piece of witness testimony is very convincing evidence
against the suspects, but that the prosecution still hesitates to use
it, since doing so would inevitably disclose the witnesses' identity,
and the mafia gang is known to be very violent against people who
testify against them. In such a case, the prosecutor will try to avoid
using this piece of evidence as long as possible, and try to win the
case in other ways. Only if these other ways fail, the prosecutor may
decide to use the witness testimony.

The parallel between this criminal court case and a 
\texttt{type-2-}argumentation
framework is easy to do: the pieces of evidence which can be disclosed are the
unrestricted arguments, and the witness testimony is a restricted argument.
\footnote{It is this example, suggested by the reviewers, which prompted to us
the expression ``restricted argument'', in the sense of 
``confidential argument''.}

The set of arguments advanced by the prosecution is a conflict-free 
set, which defends
all its elements and which contains the restricted argument only if 
it defends a piece
of evidence which could not be defended otherwise. One can imagine 
that the set of pieces
of evidence which can be disclosed is maximal, since the more pieces 
of evidence
are justified, the better the suspects' guilt is ensured. Therefore, 
the set of arguments
advanced by the prosecution is a min-def extension.


\subsection{CONFIGURATION OF A WARPLANE}

The following illustration shows that our work can be applied to non
dialogical contexts, to a configuration problem for instance.

Assume that the army wants to configure its warplanes in order for 
them to defend
against any attack of the enemy during the next mission. Among the
functions which can be put on a plane, some are desirable and even
essential, like flight, propulsion and communication functions; others
are optional, like the jammers of radar. A maximum of desirable
functions must be put on a plane, whereas a minimum of optional
functions must be installed because of cost, speed and weight
constraints. The enemy has materials which threaten some functions;
for instance, their jammer of flight threatens the flight function.

A parallel can easily be established between this problem of
configuration and a {\tt type-2-}partitioned argumentation framework:
we set $ A $ to be the set of functions and threats; $ F $ is then
the set of functions, and $ F_u $ is the set of desirable functions
whereas $ F_r $ is the set of optional functions; the attack relation
$ R $ corresponds to the conflicts between
functions and threats.

A good configuration of a plane is defined as a set of functions $C$
such that:
\begin{enumerate}
\item $C$ is conflict-free;
\item if a function in $C$ is
threatened, then $C$ contains a function which attacks the threat;
\item $C$ contains a maximum of desirable functions and a minimum of
optional functions.
\end{enumerate}

The parallel established with a {\tt type-2-}partitioned argumentation
framework shows that a good configuration is a set included in $F$
which is conflict-free (1) and which defends all its arguments (2);
therefore, it is an admissible set included in $F$. This set contains
a maximum of unrestricted arguments (the desirable functions), and a minimum
of restricted arguments (the optional functions)
(3); in other words, this admissible set is
$\prec$-maximal. Consequently, according to
property~\ref{pro_resultat1}, a good configuration is a min-def
extension.


\section{CONCLUSION AND FUTURE WORKS} \label{algorithms}

In this paper, using a partition of the set of arguments, we propose
some refinements of the argumentation framework defined
by~\cite{Dun95}.
The aim of this partition is to privilege one subset of arguments and, in
addition, to make a distinction between restricted and unrestricted arguments.
In this context, we have
defined different semantics and the associated acceptable sets:
\begin{itemize}
\item the \emph{\doadmissible{} sets} are admissible sets of arguments
   in which the presence of
   each restricted argument is justified (a restricted argument belongs to a
   \doadmissible{} set only if it defends an unrestricted argument);
\item the \emph{min-def extensions} are \doadmissible{} sets which
   respect the following condition: they maximize the unrestricted
   arguments and minimize the restricted arguments.
\end{itemize}

Different properties are presented which show the relations between the
different semantics.

Some of these results (reported in section~\ref{formalization}) 
suggest that the min-def
extensions can be computed from the set of the preferred extensions
on $F$.

Actually, from property~\ref{pro_consequence1},  it is clear that 
each min-def extension can be
obtained from one of the preferred extensions on $F$,  by minimizing its
intersection with $F_r$ (its restricted part).  It follows from
property~\ref{pro_consequence2}
that it is sufficient to consider the preferred extensions on $F$ such
that their intersection with $F_u$ is maximal (for set-inclusion).

\cite{DM01b} proposes an algorithm based on the technique of set-enumeration
for computing all the preferred extensions of a given argumentation
framework.  The basic idea is to reduce the number of generated sets
of arguments using properties of the preferred extensions.  That
algorithm can be adapted  using the following additional properties:
we are interested in subsets of $F$,  which are preferred extensions on
$F$ and whose unrestricted part (intersection with $F_u$) is maximal for
set-inclusion.

So,  we propose a two-steps method for computing min-def extensions:

\begin{enumerate}
\item  Compute $X_1$,  $X_2$,  \ldots,  $X_n$ the preferred 
extensions on $F$ with a
   maximal unrestricted part.
\item From each $X_i$,  compute the admissible sets
   by removing as many elements of $X_{ir}$ ({\it i.e.}  restricted elements) as
   possible.
\end{enumerate}
The above method is correct and complete (see~\cite{CDLSM02}).

In the particular case where there is no cycle in $F$,  the first step
in the above algorithm produces only one preferred extension on $F$.

\medskip

We consider  an extension of this model to the case where the set
of restricted arguments would be stratified. This is motivated by
several examples. For instance, among the optional functions which can
be put on a plane, some can be more costly than others; one wants to
choose the least expensive in priority. Moreover, we have seen that the
restricted arguments of an agent can be those which contain personal
information. These informations can be more or less personal, and then
one prefers to disclose the arguments which contain the least personal
information in priority. Consequently, we want to extend a {\tt
   type-2-}partitioned argumentation framework with a priority relation
over the restricted arguments. We will integrate this priority in the
definition of the $\prec$-relation and of the min-def semantics in
order to choose the restricted arguments which have the greatest
priority ({\it i.e.} those
which contain the least expensive functions or the least personal
information) for the defence of the unrestricted arguments.


\subsubsection*{Acknowledgements}
We would like to thank Hélène Fargier for fruitful preliminary 
discussions on this topic.
We would also like to thank the referees for helpful comments; in 
particular, they
suggested the ``mafia example''.


\bibliographystyle{apalike}
\bibliography{mcl}

\end{document}